\documentclass[final]{cvpr}

\usepackage{times}
\usepackage{epsfig}
\usepackage{graphicx}
\usepackage{amsmath}
\usepackage{amssymb}
\usepackage{xcolor}
\usepackage{subfig}
\usepackage{multirow}
\usepackage[ruled]{algorithm2e}
\usepackage{setspace}
\usepackage{algpseudocode}

\usepackage[pagebackref=true,breaklinks=true,colorlinks,bookmarks=false]{hyperref}

\def\cvprPaperID{****} 
\def\confYear{CVPR 2021}

\begin{document}

\title{CRD-CGAN: Category-Consistent and Relativistic Constraints for Diverse Text-to-Image Generation}

\author{Tao Hu\textsuperscript{1} ~~~~~~~~~~~~~
Chengjiang Long\textsuperscript{2}\thanks{This work was co-supervised by Chengjiang Long and Chunxia Xiao. Chunxia Xiao is the corresponding author.} ~~~~~~~~~~~~~~
Chunxia~Xiao\textsuperscript{3}${}^\ast$\\
\textsuperscript{1}{School of Information Engineering, Hubei Minzu University, Enshi, Hubei, China 445000} \\
\textsuperscript{2}{JD Finance America Corporation, Mountain View, CA, USA 94043} \\
\textsuperscript{3}{School of Computer Science,  Wuhan University, Wuhan, Hubei, China 430072} \\
{\tt\small hutao\_es@foxmail.com, cjfykx@gmail.com, cxxiao@whu.edu.cn}
}


\maketitle

\begin{abstract}
   Generating photo-realistic images from a text description is a challenging problem in computer vision. Previous works have shown promising performance to generate synthetic images conditional on text by Generative Adversarial Networks (GANs). In this paper, we focus on the category-consistent and relativistic diverse constraints to optimize the diversity of synthetic images. Based on those constraints, a category-consistent and relativistic diverse conditional GAN (CRD-CGAN) is proposed to synthesize $K$ photo-realistic images simultaneously. We use the attention loss and diversity loss to improve the sensitivity of the GAN to word attention and noises. Then, we employ the relativistic conditional loss to estimate the probability of relatively real or fake for synthetic images, which can improve the performance of basic conditional loss. Finally, we introduce a category-consistent loss to alleviate the over-category issues between $K$ synthetic images. We evaluate our approach using the Birds-200-2011, Oxford-102 flower and MSCOCO 2014 datasets, and the extensive experiments demonstrate superiority of the proposed method in comparison with state-of-the-art methods in terms of photorealistic and diversity of the generated synthetic images. 
\end{abstract}


\section{Introduction}

Text-to-image generation has wide range of applications in computer vision and graphics~\cite{Hu2021VRT, Long:WACV2019, Long:ICCV2013A, Hua:ICCV2013B, Long:ICCV2015, Long:IJCV2016, Long:CVPR2017, Hua:TPAMI2018}, and many methods have been proposed for this research topic. Various conditional Generative Adversarial Networks (GANs)~\cite{goodfellow2014generative,mirza2014conditional,reed2016learning,reed2016generative,ledig2017photo,stackgan,StackGANplus,zhang2018self,Xu:CVPR2018,Mao_2019_CVPR,Yin_2019_CVPR,ChaGK19,tan2019text2scene,li2019storygan,li2019object-driven,eghbal2019mixture,cheng2020rifegan,liang2019cpgan,Hu2021VRT,koh2021text,GAO2021107384,yang2021multi} have been developed to generate photo-realistic images conditional on text with a random noise. However, it is still challenging to simultaneously generate a set of photo-realistic as well as significantly diverse synthetic images conditional on text description.

Existing text-to-image GANs mainly focus on improve the synthesizing performance to generate high-quality and resolution images by tree-liked stacked GANs~\cite{stackgan,StackGANplus}, word-region attention guided GANs~\cite{Xu:CVPR2018,zhang2018self,park2019semantic}, object-driven attentive GANs~\cite{li2019object-driven}, or the mode seeking GANs~\cite{Mao_2019_CVPR,Hu2020HierarchicalME}. However, all those methods ignore the category attributes of the real image corresponding to the text description. Such as Liu \emph{et al.}~\cite{Liu2020} used the category information to generate text. It means that we expect the synthetic images generated by GANs should have category attributes corresponding to real images. In other words, we expect the synthetic images have the main visual feature of the same category. 

\begin{figure*}[h]
\begin{center}
\includegraphics[height=0.24\linewidth, width=0.95\linewidth, angle=0]{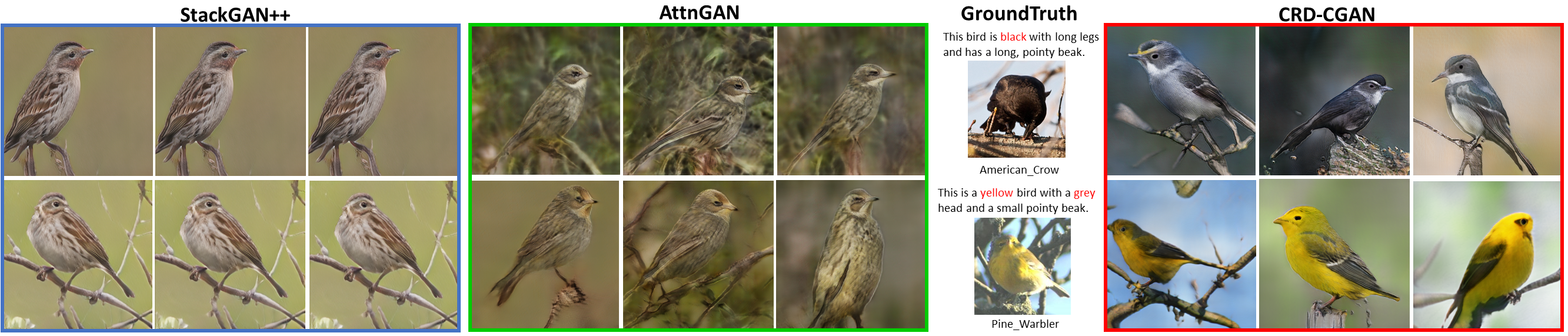}
\end{center}
\caption{Illustration of three methods to generate a synthetic images conditional on a text. Our goal is to generate a set of diverse and high-quality synthetic images that are as consistent as possible with the text description and consistent with the category visual feature of real image category.}
\label{fig:introduction}
\end{figure*}

When given a text description (middle), the synthetic images generated by StackGAN++~\cite{StackGANplus} (in the blue rectangle in Figure~\ref{fig:introduction}) are highly realistic. But there is a clear visual difference between synthetic and real images. The color of {\em ``American\_Crow"} bird category is black, but the color of synthetic bird generated by StackGAN++ is grey. The same problem appears on the {\em ``Pine\_Warbler"} category. The synthetic images generated by AttnGAN~\cite{Xu:CVPR2018} also have not consistent of color with the real image, which are shown in green rectangle in Figure~\ref{fig:introduction}. And there are less diversity between them. Therefore, it is desirable to ensure the synthetic images to retain the main visual feature of real image, as well as preserving the category-consistent visual concept and keeping diversity.

To address this issue, we propose a novel diverse photo-realistic and category-consistency text-to-image generation method that effectively exploits the relative relationship between synthetic and real images, and the category information of the real images correspondingly within the generation procedure, named as category-consistent and relativistic diverse conditional GAN (CRD-CGAN). Inspired by the advantages of tree-liked stacked GANs~\cite{huang2017stacked,stackgan,StackGANplus,Xu:CVPR2018,Mao_2019_CVPR,Hu2021VRT}, we propose a basic diverse conditional GAN (D-CGAN) to generate $K$ synthetic images with $K$ different generators and one shared discriminator firstly. Then we propose a relativistic discrimination regularization to improve the estimation performance of synthetic image is real, which can effectively generate more diverse $K$ photo-realistic images. To ensure that the $K$ synthetic images have the main visual feature of category correspondingly, we use the category consistency regularization to constrain the main visual feature of synthetic images. From Figure~\ref{fig:introduction}, we can see the synthetic images generated by CRD-CGAN retain the main visual feature of real images with a diversity of photo-realistic appearance.

To sum up, our contributions are three-fold as follows:

(1) We propose a new framework CRD-CGAN which contains $K$ generators and a shared discriminator to improve the diversity of $K$ high-quality synthetic images simultaneously.  

(2) We incorporate category-consistent and relativistic diverse conditional constraints, which effectively improve the quality of photo-realistic synthetic images and ensure that $K$ synthetic images retain the main visual feature of the corresponding category of real images.

(3) The proposed CRD-CGAN achieves the state-of-the-art performance on the Caltech-UCSD Birds-200-2011 dataset~\cite{Wah:CUB_200_2011}, the Oxford 102 Category Flower dataset~\cite{Nilsback:2008automated} and the MS COCO 2014 dataset~\cite{lin2014microsoft} for text-to-image generation.

\section{Related Works}
Generative Adversarial Networks (GANs) and attention mechanisms have been successfully applied to various visual applications~\cite{Ding:ICCV2019, Wei:CGF2019, Zhang:AAAI2020, Liu:CVPR2020, Islam:CVPR2020, Zhang:CGF2020, Zhang:ICME2020, Vasu:WACV2020, Zhang:TCSVT2021, Islam:AAAI2021}. Especially, to translate the visual concepts from characters to pixels, Reed \emph{et al.}~\cite{reed2016learning} designed a novel GAN to effectively bridge text and image modeling, while the size of synthetic image is $64 \times 64$ pixels. At the same time, Reed \emph{et al.}~\cite{reed2016generative} used the generative adversarial what-where network to estimate what content to draw in which location and generated $64 \times 64$ synthetic image conditioned on text. Due to the real image distribution and GAN's distibution may not overlap in high dimensional pixel space, Zhang \emph{et al.}~\cite{stackgan} proposed StackGAN to generate $256 \times 256$ photo-realistic image conditioned on text in two separate stages. However, the StackGAN is not stable enough in GAN's training. Zhang \emph{et al.}~\cite{StackGANplus} extended StacKGAN to StackGAN++ to improve the quality of generated images by jointly approximating multiple distributions, while the multiple generators and discriminators arranged in a tree-like structure. Xu \emph{et al.}~\cite{Xu:CVPR2018} proposed an AttnGAN to synthesize fine-grained details at different sub-regions of the generated image by automatically attending to the relevant words. Li \emph{et al.}~\cite{li2019object-driven} proposed object-driven attentive GANs to synetesize objects by paying attention to the most relevant words and the pre-generated layout. Hu \emph{et al.}~\cite{Hu2021VRT} proposed a diverse GAN-Visual Representation on Text, which extracts visual features from synthetic images generated by a DCGAN, for visual recognition. The above-mentioned related work are focused on generating higher resolution and realistic images conditioned on text. In contrast, our work is going to synchronously generate more diverse and photo-realistic synthetic images with the same resolution.

Some state-of-the-art GANs are designed to generate significantly diverse synthetic images by optimizing the generator or discriminators. For example, Mao \emph{et al.}~\cite{Mao_2019_CVPR} proposed a mode seeking regularization method to minimize the generator's loss by maximize the ratio between the distance of two synthetic images in visual space and the distance of two noises correspondingly, and they used different noise inputs to generate diverse synthetic images through the optimized generator. To generate more diverse images, Cha \emph{et al.}~\cite{ChaGK19} used triplets ({\em i.e.}, a positive image, a text, and a negative-image) to train the generator and discriminator, and selected the negative image by the semantic distance from a positive example in the class. And Hu \emph{et al.}~\cite{Hu2020HierarchicalME} used the hierarchical model in conditional GANs to imporve the diversity of synthetic images. Wang \emph{et al.}~\cite{Wang2020StochasticCG} proposed a stochastic conditional GNAs to generated diverse synthetic images with the same input condition. StarGAN v2~\cite{choi2020stargan} learns a specific style code from a given reference image to address the style diversity, but it did not be extended to text-to-image fields.

\begin{figure*}[h]
\begin{center}
\includegraphics[height=0.28\linewidth, width=0.99\linewidth, angle=0]{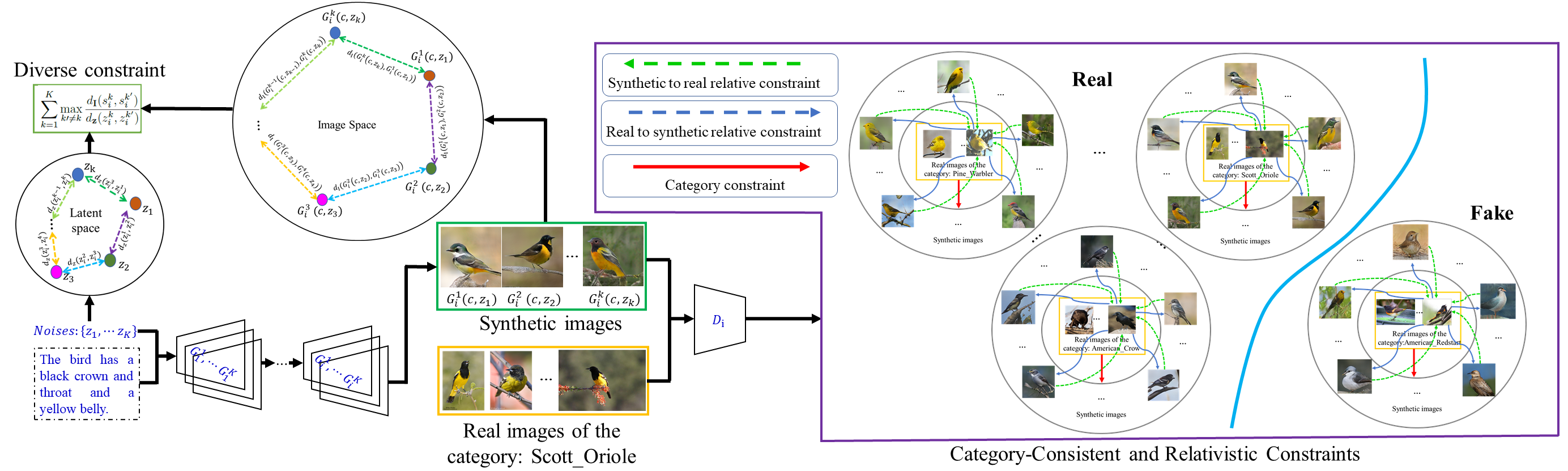}
\end{center}
\caption{Overview of our proposed framework CRD-CGAN. We use the generators $G_{i}^1,...,G_{i}^k$ to generate $K$ synthetic images. Based on the difference between image space and latent noise space, we use the diverse constraints to improve the diversity of $K$ synthetic images. The proposed category-consistent and relativistic constraints are reflected in the discriminator at each stage. In this figure, the green arrow is the relativistic loss between synthetic image with the corresponding real images based on the true label, and the blue arrow is the relativistic loss between real image with the corresponding synthetic images based on the fake label. Among these four groups of images (inner circle are real images and outsider are the corresponding synthetic images), the three ground above the decision boundary have better category consistency.}
\label{fig:framework}
\end{figure*}

It is well known that the real image corresponding to the text has very specific category information. For example, each image in the Birds-200-2011 dataset~\cite{Wah:CUB_200_2011} has a specific bird category, such as ``American\_Crow", and ``Scott\_Oriole". However, all those methods ignore the category attributes of the real image corresponding to the text description. The category attributes can effectively improve the GAN's performance, for example, Liu \emph{et al.}~\cite{Liu2020} used the category information to generate text. It means that we expect the synthetic images generated by GANs should have the same category attributes as the corresponding real images. In other words, we expect the synthetic images have the main visual feature of the same category. 

Unlike previous approaches that generate more realistic or diverse images by cyclic inputing set of noise, our proposed method extend one generator to $K$ generators in a certain resolution. In this paper, we propose a novel GAN framework named CRD-CGAN which incorporates category-consistent and relativistic constraints for diverse image generation.
It exploits the relative relationship between synthetic and real images, and the category information of the real images correspondingly in the generation procedure. Inspired by the advantages of tree-liked stacked GANs~\cite{huang2017stacked,stackgan,StackGANplus,Xu:CVPR2018,Mao_2019_CVPR}, we first design a basic diverse conditional GAN (D-CGAN) to generate $K$ synthetic images with $K$ different generators and one shared discriminator. Then we propose a relativistic discrimination constraint to improve the photo-realistic performance of synthetic images. To ensure that the $K$ synthetic images have the main visual feature of category correspondingly, we use the category consistency regularization to constrain the main visual features of synthetic images. From Figure~\ref{fig:introduction}, we can see the synthetic images generated by CRD-CGAN retain the main visual appearance feature of real images with a diversity of photo-realistic.

\section{Proposed Method}

Our CRD-CGAN consists of two key components,  {\em i.e.}, the diverse conditional GAN (D-CGAN) with attention diverse constraint. The CRD-CGAN with the discrimination regularization in combination of relativistic conditional constraint and category consistent constraint. The D-CGAN is the standard network for generating $K$ diversity synthetic images. The CRD-CGAN use relativistic conditional regularization and category consistency regularization to improve the quality and diversity of synthetic images. As illustrated in Figure~\ref{fig:framework}, with $K$ generators at the $i$-th stage, $K$ synthetic images have been generated by our CRD-CGAN. 

The $K$ synthetic images are feed into the generators and discriminator optimization process. We firstly describe a diverse conditional GAN which is denoted as D-CGAN for diverse text-to-image generation, and then employ category-consistent and relativistic constraints to improve the quality and diversity of synthetic images.

\subsection{Diverse Text-to-Image Generation}

The D-CGAN is uesd to generate $K$ photo-realistic synthetic images simultaneously with $K$ generators and one shared discriminator. The $K$ generators $\left \{ G_{1},G_{2},...,G_{K} \right \}$ use $K$ different prior noise vectors $\left \{ z_{1},z_{2},...,z_{K} \right \}$ to ensure the $K$ synthetic images have high diversity, correspondingly. The discriminator $D$ and generators $G_1, \ldots, G_K$ can be optimized in a joint form by alternatively maximizing $\mathcal{L}_{D}$ and minimizing $\mathcal{L}_{G_1, \ldots, G_K}$ until convergence.


To make sure that the generated synthetic images cover the text description, we use image-text similarity loss, denoted as $\mathcal{L}_{sim}(s_1, \ldots, s_K)$, which is used to estimate the probability of the matching level between each word and a sub-region of synthetic image, and the matching level between the input sentence and synthetic image. Following~\cite{Xu:CVPR2018}, $\mathcal{L}_{sim}(s_1, \ldots, s_K)$ is defined as:
\begin{equation}
\begin{split}
\mathcal{L}_{sim}(s_1, \ldots, s_K)=
\sum_{k=1}^{K}(\mathcal{L}_{1,k}^{w}+\mathcal{L}_{2,k}^{w}+\mathcal{L}_{1,k}^{s}+\mathcal{L}_{2,k}^{s}) 
\label{attention-loss}
\end{split}
\end{equation}
where $\mathcal{L}_{1,k}^{w}$ is the negative log posterior probability that measure matching level between the synthetic images and the corresponding word-level description of the $k$-th generator, and $\mathcal{L}_{2,k}^{w}$ is the negative log posterior probability measure matching level between the word-level description with the corresponding image. Similarly, the $\mathcal{L}_{1,k}^{s}$ and $\mathcal{L}_{2,k}^{s}$ are the negative log posterior probabilities of matching level between image and sentence-level description of the $k$-th generator. The objective function of attention loss is to minimize $\mathcal{L}_{sim}(s_1, \ldots, s_K)$.

To ensure the diversity among the generated synthetic images, we introduce a diversity loss, denoted as $\mathcal{L}_{div}(s_1, \ldots, s_K)$, to measure the diversity of the $K$ synthetic images. Inspired by~\cite{Mao_2019_CVPR}, we define it as:
\begin{equation}
\begin{split}
\mathcal{L}_{div}(s_1, \ldots, s_K) = \sum\limits_{k=1}^K \max\limits_{k\prime \neq k} \frac{d_{\bf I}(s_k, s_{k^\prime})}{d_{\bf z}(z_k, z_{k^\prime})}\label{eqn:diverseloss}
\end{split}
\end{equation}
where $d_{\bf I}$ is the distance between synthetic image features, and $d_{\bf Z}$ means the distance between noise vector. 

At each stage $i$, the loss of generators  $\left \{ G_{1},G_{2},...,G_{K} \right \}$ and discriminator $D_{i}$ of D-CGAN can be defined as Eq.~\ref{base-generator-loss} and Eq.~\ref{base-dis-loss} generally, where $c$ is the condition parameter, $s_{i}^{k}$ is from the synthetic image distribution $p_{G_{i}^{k}}$, and $X_{i}$ is from the true image distribution $p_{data_{i}}$. The images from interpolated text embedding can fill in the gaps in the data manifold, which were presented during training. With the diverse term in Eq.~\ref{eqn:diverseloss} included, we are able to ensure the diversity of the synthetic images.
\begin{equation}
\begin{split}
\label{base-generator-loss}
\mathcal{L}^{DIV}_{G_{i}}=
\sum_{k=1}^{K}\mathbb{E}_{s_{i}^{k}\sim p_{G_{i}^{k}}}\left [ -\text{log}(D_{i}(s_{i}^{k},c) \right ]\\
+\mathcal{L}_{sim(\ldots)}+\mathcal{L}_{div(\ldots)}
\end{split}
\end{equation}

\begin{equation}
\label{base-dis-loss}
\resizebox{.88\linewidth}{!}{$
\displaystyle
\mathcal{L}^{DIV}_{D_{i}}=
K\mathbb{E}_{X_{i}\sim p_{data_{i}}}\left [ -\text{log}(D_{i}(X_{i},c)) \right ]+\\
\sum_{k=1}^{K}\mathbb{E}_{s_{i}^{k}\sim p_{G_{i}^{k}}}\left [ -\text{log}(1-D_{i}(s_{i}^{k},c))  \right ]
$}
\end{equation}

The generators $\mathcal{L}^{DIV}_{G_{i}}$ are trained to combine $K$ different prior noise vectors, and the text embedding vector is used to interpolate $K$ different synthetic images. The discriminator $\mathcal{L}^{DIV}_{D_{i}}$ has been trained to predict whether the synthetic images and the text match or not. 

\begin{figure*}[t]
\begin{center}
\includegraphics[height=0.280\linewidth, width=0.998\linewidth, angle=0]{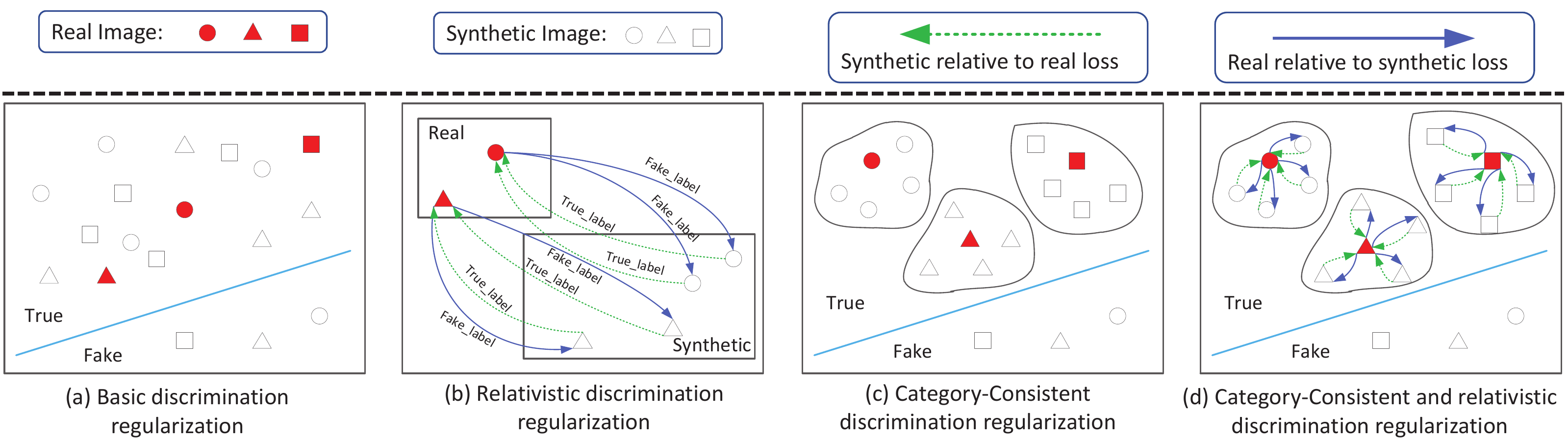}
\end{center}
\caption{Illustration of four different kinds of discrimination regularization. (a) is the basic discrimination regularization, which just discriminates whether the synthetic image is true. (b) is the proposed relativistic discrimination regularization, which adds the relativistic average conditional loss on the basis of (a). (c) is the proposed category-consistent discrimination regularization, which combines the category consistency loss on the basis of (a). (d) is the proposed category-consistent and relativistic discrimination regularization, which introduce the category consistency loss and the relativistic average conditional loss on the basis of (a). The green arrow is the relativistic loss from synthetic image to the corresponding real images based on the true label. The blue arrow is the relativistic loss from real image to the corresponding synthetic images based on the fake label.}
\label{fig:regular}
\end{figure*}

\subsection{Category-Consistent and Relativistic Constraints}

To better explore the visual feature representation of text, we design CRD-CGAN to generate $K$ diverse synthetic images by alternative optimizing generators and discriminators with category-consistent and relativistic constraints. And Based on the two constraints, we proposed the category-consistent and relativistic discrimination regularization. To understand the meaning of a specific discrimination regularization, we discuss each regularization and the corresponding variant of CRD-CGAN as follow:

(a) {\textbf{Basic discrimination regularization}}(Figure~\ref{fig:regular}(a)) with the standard conditional loss is used to estimate the probability that the synthetic image is real in the variant D-CGAN.

(b) {\textbf{Relativistic discrimination regularization}}(Figure~\ref{fig:regular}(b)) with relativistic conditional loss can be used to estimate the probability that the synthetic image is enhanced realistic than a randomly sampled synthetic image, which can improve the performance of conditional loss. The variant with this regularization can be denoted as RD-CGAN.

(c) {\textbf{Category-consistent discrimination regularization}}(Figure~\ref{fig:regular}(c)) with category consistency loss is proposed to alleviate the over-category issue between $K$ generators based on the image category. There is a performance imbalance between $K$ generators $\left \{ G_{1},G_{2},...,G_{K} \right \}$, for example, some synthetic images is significantly different in shape or color from the corresponding real image. The corresponding variant with this regularization can be denoted as CD-CGAN.

(d) {\textbf{Category-consistent and relativistic discrimination regularization}}(Figure~\ref{fig:regular}(d)) in term of combining both a relativistic conditional loss and a category consistency loss is exploited in our proposed CRD-CGAN to jointly improve synthetic image's quality and diversity. 

\subsubsection{Relativistic conditional loss}

Since the gradient of the non-saturation loss is not zero and unstable, the discriminator in the traditional GAN cannot be optimized, or a higher learning rate is required during the training process. The relativistic conditional loss is used to estimate the probability, which is compared to a randomly sampled synthetic image, thereby improving the performance of conditional loss in the process of training the discriminator. Inspired by~\cite{jolicoeur2018rfdiv,jolicoeurmartineau2019the}, to estimate the probability if the input sample is real, we simultaneously use the real and fake data. In other words, the estimated probability from absolute real to fake relative to real or fake. The relativistic conditional discriminator ${D_{i}}$ is defined as ${\textit{sigmoid}}((D_{i}(X_r,c)-D_{i}(X_f,c))\times l)$, where $l$ is either 1 or $-1$. If $X_r$ is more realism relative to $X_f$ ($D_{i}(X_r,c)>D_{i}(X_f,c)$) or $X_f$ is more artifacts relative to $X_r$ ($D_{i}(X_f,c)>D_{i}(X_c,c)$), we set $l=1$ and $l=-1$ otherwise. 

When measuring the probability that the synthetic image is real, the ${D_{i}}$ is hard to converge. To address this issue, the relativistic discrimination regularization is proposed to find out all possible combinations of real and synthetic image in the mini-batch. It compares the critic of the real image to the average critic of synthetic images  with true or fake label correspondingly. The loss function of relativistic 
conditional generators and discriminator are defined as $\mathcal{L}^{RE}_{G_{i}}$ and $\mathcal{L}^{RE}_{D_{i}}$.

In the Eqs.(5)-(7), $l_r$ and $l_f$ are symmetric labels~\cite{mao2017least}, e.g., -1 and 1.
$R(s_{i}^{k},X_{i},D_{i},c,l_r)$ in Eqn~(\ref{relative-generator-loss}) is used to calculate the probability of more realism of $s_{i}^{k}$ relative to $X_{i}$, and $R(X_{i},s_{i}^{k},D_{i},c, l_f)$ in Eqn~(\ref{relative-generator-loss}) is used to calculate the probability of more artifacts of $X_{i}$ relative to $s_{i}^{k}$. Eqn~(\ref{relative-dis-loss}) calculates the probability more realism of  $X_{i}$ relative to $s_{i}^{k}$ and the probability of more artifacts of $s_{i}^{k}$ relative to $X_{i}$.

\begin{equation}
\resizebox{.90\linewidth}{!}{$
\displaystyle
R(X,Y,D,c,l)=\text{log}(\text{sigmoid}((D(X,c)-D(Y,c))\times l))
$}
\end{equation}
\vspace{-0.25cm}
\begin{equation}
\label{relative-generator-loss}
\resizebox{.88\linewidth}{!}{$
\displaystyle
\mathcal{L}^{RE}_{G_{i}}=
\sum_{k=1}^{K}\mathbb{E}^{s_{i}^{k}\sim p_{G_{i}^{k}}}_{X_{i}\sim p_{data_{i}}}
\left[R(s_{i}^{k},X_{i},D_{i},c, l_r)  \right ]\\+
\sum_{k=1}^{K}\mathbb{E}^{s_{i}^{k}\sim p_{G_{i}^{k}}}_{X_{i}\sim p_{data_{i}}}
\left [R(X_{i},s_{i}^{k},D_{i},c, l_f) \right ]
$}
\end{equation}
\vspace{-0.25cm}
\begin{equation}
\label{relative-dis-loss}
\resizebox{.88\linewidth}{!}{$
\displaystyle
\mathcal{L}^{RE}_{D_{i}}=
\sum_{k=1}^{K}\mathbb{E}^{s_{i}^{k}\sim p_{G_{i}^{k}}}_{X_{i}\sim p_{data_{i}}}
\left [R(X_{i},s_{i}^{k},D_{i},c,l_r) \right ]\\+
\sum_{k=1}^{K}\mathbb{E}^{s_{i}^{k}\sim p_{G_{i}^{k}}}_{X_{i}\sim p_{data_{i}}}
\left[R(s_{i}^{k},X_{i},D_{i},c,l_f)  \right ]
$}
\end{equation}

\subsubsection{Category consistency loss}

Considering that each category of images has its unique features, such as the shape and color of objects. The conditional loss just estimates that the probability of synthetic image is real without the category attributes. To further improve the performance of D-CGAN, we propose the category consistency loss $\mathcal{L}_{G_i}^{CC}$. It is used to estimate the probability that the synthetic images belong to the same category of the corresponding real image.

After extracting the visual features of synthetic images and real image by the image encoder, we use a softmax layer~\cite{krizhevsky2012imagenet} to infer the probability distributions of each visual feature. We can use the cross-entropy to estimate the probability that the real image belongs to the corresponding category, but estimating the category performance of synthetic images is very difficult using the same method directly. This is because compared to real images, the synthesized image generated by D-CGAN does not have all visual features. 

We extract the real image feature $X_{i}$ and synthetic image feature $s_{i}^{k}$ using the same image encoder. To correlate the relationship between synthetic images and categories, we calculate the cosine similarity~\cite{Cosine2019} $\textit{Sim}(X_{i},s_{i}^{k})$ between real image feature $X_{i}$ and synthetic image feature $s_{i}^{k}$. The $\textit{Sim}(X_{i},s_{i}^{k})$ is defined as the correlation weight. Based on the correlation weight, we concatenate the real image feature $X_{i}$ and synthetic image feature $s_{i}^{k}$ as 
the mixed feature. We apply a linear classification to produce classification score before softmax layer. Finally, we use the softmax layer to yield the category probability of the combined visual feature. If the generated synthetic images have the same category, the combined features should enforce the correct classification. Otherwise, the combined visual feature might weaken the classification confidence and even lead to misclassification. Therefore, we define the category consistency loss as a cross-entropy between prediction probabilities and the true category $c_i$:
\begin{equation}
\begin{split}
\mathcal{L}_{G_i}^{CC}=-\sum_{y=1}^{Y}\delta(y=c_i)\textit{log}P{(y|X_i,s_i^1, \ldots, s_i^K)} 
\label{category-loss}
\end{split}
\end{equation}
where $Y$ is the total category number of input dataset, $\delta(y=c_i)$ is 1 when $y$ is the true category $c_i$ of $X_i$ and 0 otherwise,  and $P{(y|X_i,s_i^1, \ldots, s_i^K)}$ is defined as follow:
\begin{equation}
\begin{split}
&P(y|X_i, s_i^1, \ldots, s_i^K)\\
&=
\textit{softmax} \left( {\bf W}^T \left [X_i,\lambda \sum_{k=1}^{K}s_{i}^{k}\otimes \textit{Sim}({X_{i},s_{i}^{k}}) \right] + {\bf b} \right)
\end{split}
\label{pro-discr-combine-feature}
\end{equation}
where $\lambda$ is the adjustment factor to balance the importance of synthetic image feature $s_{i}^{k}$, ${\bf W}$ and ${\bf b}$ are the linear classification parameters. The $\mathcal{L}_{G_i}^{CC}$ estimates the maximum likelihood that the synthetic images and real images belongs to the same category. In other words, it can improve the semantic consistency of synthetic images compared with the corresponding real image. 

Based on the Eqs.~\ref{relative-generator-loss}, \ref{relative-dis-loss} and~\ref{category-loss}, the final loss function for generator and discriminator in our CRD-CGAN can be formulated as follows:
\begin{equation}
\begin{split}
\mathcal{L}_{G_{i}}=\mathcal{L}^{DIV}_{G_{i}}+\mathcal{L}^{RE}_{G_{i}}+\delta \mathcal{L}_{G_i}^{CC}
\end{split}
\end{equation}
\begin{equation}
\mathcal{L}_{D_{i}}=\mathcal{L}^{DIV}_{D_{i}}+\mathcal{L}^{RE}_{D_{i}}
\end{equation}
where $\delta$ is the weight of category consistent constraint. We set $\delta=1$ in the following experiments.

\section{Experiments}

\subsection{Experiment Settings}

\textbf{Datasets}. In this paper, we evaluate our method on the Caltech-UCSD Birds-200-2011 Dataset~\cite{Wah:CUB_200_2011} and the Oxford 102 Category Flower Dataset~\cite{Nilsback:2008automated}. To evaluate our method in the multi-objects complex scenes, we evaluate our method on the MS COCO 2014 Dataset~\cite{lin2014microsoft}.

The Birds-200-2011 Dataset consists of 11,169 bird images from 200 categories and each category has 60 images averagely. We randomly select 9,935 images for training, and use the rest 1,234 images for testing. The dataset is very challenging because it contains images with multiple objects and various backgrounds. The Oxford-102 flower Dataset~\cite{Nilsback:2008automated} consists of 8,189 images with 102 categories of flowers which commonly occurs in the United Kingdom, and each category has 40 to 258 images. The MS COCO 2014 Dataset~\cite{lin2014microsoft} contains images of 91 object categories, which contains 82783 training images, 40504 validation images and 40775 testing images. 

\textbf{Evaluation Metrics}. We use Fréchet Inception Distance~\cite{heusel2017gans}, denoted as FID, to evaluate the quality of synthetic images by calculating the distance between synthetic and real images through features extracted by Inception Network. Lower FID value indicates better quality of the synthetic image. We also apply the Learned Perceptual Image Patch Similarity~\cite{zhang2018unreasonable}, denoted as LPIPS, to measure performance of diverse of GANs. Higher LPIPS value means more diverse of synthetic image. The Inception score~\cite{SalimansGZCRCC16} is also used to evaluate the synthetic quality of images. The R-Precision~\cite{Xu:CVPR2018} is used to evaluate the correspondence between input text and the synthetic image.

Due to there is no category infomation in the process of calculating the FID value, we conduct a user-study on the testing datasets to further evaluate the similarity between synthetic images and corresponding real image. We first use our methods and baselines to generate 200 synthetic images each, which are based on the same 40 randomly selected texts from the validation set from each dataset. We invite 100 random volunteers to attend this experiment. The volunteer choose the images most similar to the corresponding real image, and them vote on the chose images for similarity. Based on the scoring and voting table of user-study, we use the equation~\ref{similar-score} to calculate the total average similarity $Score_{similarity}$ between the selected synthetic images and the corresponding real image with each dataset for all compare methods.

\begin{equation}
\resizebox{.90\linewidth}{!}{$
\displaystyle
socre_{similarity}=\left \{ \sum_{V=1}^{100}\left [\sum_{sentence=1}^{40}(m/5)*\sigma   \right ]/40 \right \}/100
$}
\label{similar-score}
\end{equation}
where $V$ is the number of volunteer, $sentence$ is the number of chose sentence from validation set, and $5$ means we generate 5 synthetic images based on each method.


\textbf{Baselines}. We evaluate our proposed method CRD-CGAN with StackGAN++~\cite{StackGANplus}, AttnGAN~\cite{Xu:CVPR2018}, MSGAN~\cite{Mao_2019_CVPR}, as well as the variants D-CGAN, RD-CGAN, and CD-CGAN mentioned in Section 2.2. Note that we implement two versions of D-CGAN, denoted as ``D-CGAN-A" incorporating AttnGAN and ``D-CGAN-S" incorporating StackGAN++, where D-CGAN-A or D-CGAN-S share the same structure for each generator and discriminator with AttnGAN or StackGAN++.

\begin{figure*}[h]
\begin{center}
\includegraphics[height=0.885\linewidth, width=0.998\linewidth, angle=0]{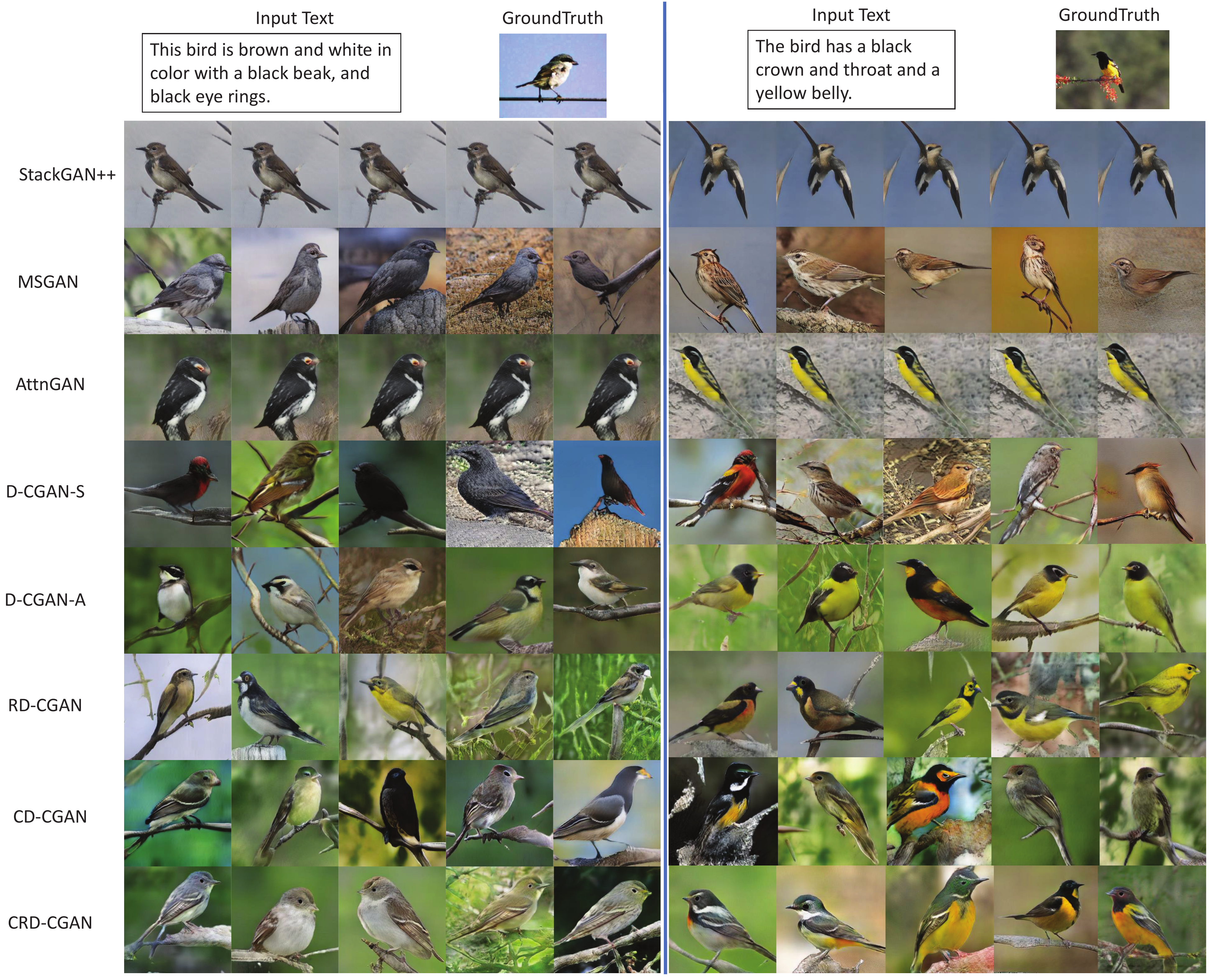}
\end{center}
\caption{Visualization of $K=5$ high-resolution and photo-realistic synthetic images conditioned on a text, and comparison with state-of-the-art methods (top) on the Birds-200-2011 dataset. }
\label{fig:birds-visual-compare}
\end{figure*}

\subsection{Comparison with the State-of-the-arts}

The FID and LPIPS scores for our proposed CRDCGAN and other methods on the Birds-200-2011 dataset are summarized in Table~\ref{tab:diverse_bird}. From Table~\ref{tab:diverse_bird}, we can see the synthetic images generated by D-CGAN-S are more diverse than by StackGAN++ and MSGAN, and the synthetic images generated by DC-GAN-A is more diverse than by AttnGAN. We also see that the synthetic images generated by CRD-CGAN are more category similarity than other methods. The synthetic images generated by CRD-CGAN exhibits the highest quality compared with real image correspondingly, which has the lowest FID score. In addition, the synthetic images generated by CRD-CGAN have the highest LPIPS score, which also shows the most diverse than other methods.

\begin{table}[h]	
\centering
\caption{Diversity performance comparison on the Birds-200-2011 dataset.} 
\resizebox{.49\textwidth}{!}{
\begin{tabular}{l|c|c|c}
\hline
Methods    & FID                   & LPIPS &  User study   \\ \hline
StackGAN++
&  
27.90$\pm $0.02     & 31.37\%$\pm $3.17    &12.17\%$\pm $1.05          \\
MSGAN
&    27.48$\pm $0.38     & 36.87\%$\pm $0.68&   15.07\%$\pm $5.62  \\
AttnGAN
& 23.81$\pm $0.53     & 35.26\%$\pm $0.05&  12.38\%$\pm $3.07   \\ \hline
D-CGAN-S  &  26.41$\pm $0.48     & 37.12\%$\pm $1.97 & 19.72\%$\pm $3.20 \\
D-CGAN-A & 22.61$\pm $0.13 $\downarrow$    & 38.67\%$\pm $0.49  &26.49\%$\pm $4.31  \\
RD-CGAN & 26.53$\pm $0.253     & 39.07\%$\pm $0.31 & 24.73\%$\pm $4.38  \\
CD-CGAN & 28.25$\pm $0.16    & 39.12\%$\pm $0.28 &  31.67\%$\pm $2.93 \\
CRD-CGAN &24.59$\pm $0.35 &39.91\%$\pm $0.34$\uparrow$&  33.24\%$\pm $5.47$\uparrow$ \\
\hline
\end{tabular}
\label{tab:diverse_bird}
}
\end{table}

We visualize some synthetic images by our CRD-CGAN and other methods in Figure~\ref{fig:birds-visual-compare}. We can observe the synthetic images generated by StackGAN++ and AttnGAN are less diverse, and the synthetic images generated by MSGAN and D-CGAN-S have lower similarity to real images while having good diversity. The synthetic images generated by our CRD-CGAN have highest similarity to real images with highest diversity. For example, the input text is {\em``This bird is brown and white in color with a black beak, and black eye rings"}. The words {\em``brown"} and {\em``white"} are the main attributes of the bird, which are reflected in the synthetic images generated by our CRD-CGAN. In addition, those synthetic images are also drawn with {\em``black beak"} and {\em``black eye rings"}.

\begin{figure*}[h]
\begin{center}
\includegraphics[height=0.885\linewidth, width=0.998\linewidth, angle=0]{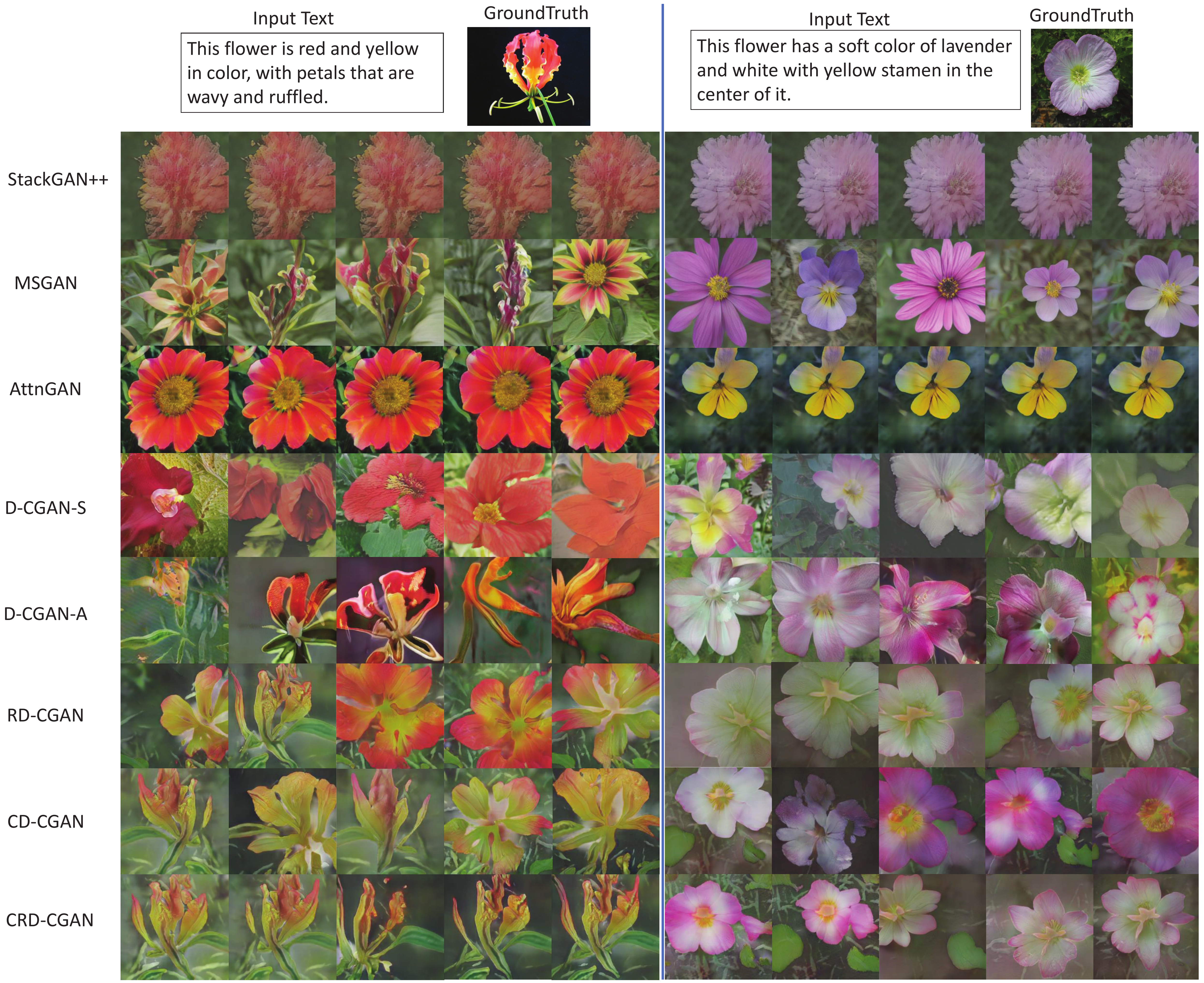}
\end{center}
\caption{Visualization of $K=5$ high-resolution and photo-realistic synthtic images condtioned on a text, and compared with the corresponding real images (top) on the Oxford 102 Flower dataset.}
\label{fig:flowers-visual-compare}
\end{figure*}

We also evaluate the diversity performance of our proposed CRD-CGAN on the Oxford-102 flower dataset with FID and LPIPS metrics in Table~\ref{tab:diverse_flowers}. From Table~\ref{tab:diverse_flowers}, we can observe that (1) our proposed D-CGAN-S, D-CGAN-A and CRD-CGAN can effectively reduce the FID value, which means the synthetic images generated by our methods have higher category consistency with the corresponding real images; (2) the synthetic images generated by our CRD-CGAN have the highest diversity than other methods. Again, the category consistency loss and relativistic conditional loss can improve the diversity of synthetic images, while the category consistency effectively constrains the quality of synthetic images; and (3) CRD-CGAN achieves the best score in user study of similarity comparison, which means the synthetic images generated by CRD-CGAN have the highest shape and color consistency with the corresponding real images.

\begin{table}[h]
\centering
\caption{Diverse Performance comparison on the Oxford-102 flower dataset.} 
\resizebox{.49\textwidth}{!}{
\begin{tabular}{l|c|c|c}
\hline
Methods    & FID                   & LPIPS       &   User study     \\ \hline
StackGAN++& 64.13$\pm $0.88     & 23.47\%$\pm $1.63 &7.47\%$\pm $0.92
\\
MSGAN& 61.95$\pm $0.23     & 32.09\%$\pm $0.29 &16.07\%$\pm $4.31 \\
AttnGAN
& 42.41$\pm $0.19     & 33.04\%$\pm $0.84 & 16.53\%$\pm $1.90 \\ \hline
D-CGAN-S  &  45.03$\pm $1.07    & 33.50\%$\pm $0.13 &19.36\%$\pm $1.41 \\
D-CGAN-A & 33.11$\pm $0.11 $\downarrow$     & 33.16\%$\pm $0.81  &22.51\%$\pm $4.56 \\
RD-CGAN & 42.76$\pm $0.23   & 33.31\%$\pm $0.80& 23.69\%$\pm $3.57  \\
CD-CGAN & 43.71$\pm $0.19 & 34.82\%$\pm $0.79 $\uparrow$& 30.11\%$\pm $3.14  \\
CRD-CGAN &40.75$\pm $0.32    &37.56\%$\pm $0.15 $\uparrow$&37.38\%$\pm $3.07$\uparrow$ \\
\hline
\end{tabular}
\label{tab:diverse_flowers}
}
\end{table}

To better understand the effectiveness of our proposed CRD-CGAN, we also visualize the generated results of CRD-CGAN and its variants on the 102 Flower dataset. As shown in Figure~\ref{fig:flowers-visual-compare}, the StackGAN++ just generates rough shape of flower. The MSGAN can keep a good quality and diversity of synthetic images, but it can not guarantee consistency with real images. Our D-CGAN-S can generate more diverse synthetic images, while it has lower image quality than D-CGAN-A and CRD-CGAN. The D-CGAN-A and CRD-CGAN can achieve the better diversity with higher image quality. 

\subsection{Analysis and Discussion}

In this section, we analyze and discuss previous experimental results to further illustrate the advantages of our proposed CRD-CGAN.

\begin{figure}[h]
\begin{center}
\includegraphics[height=0.85\linewidth, width=0.90\linewidth, angle=0]{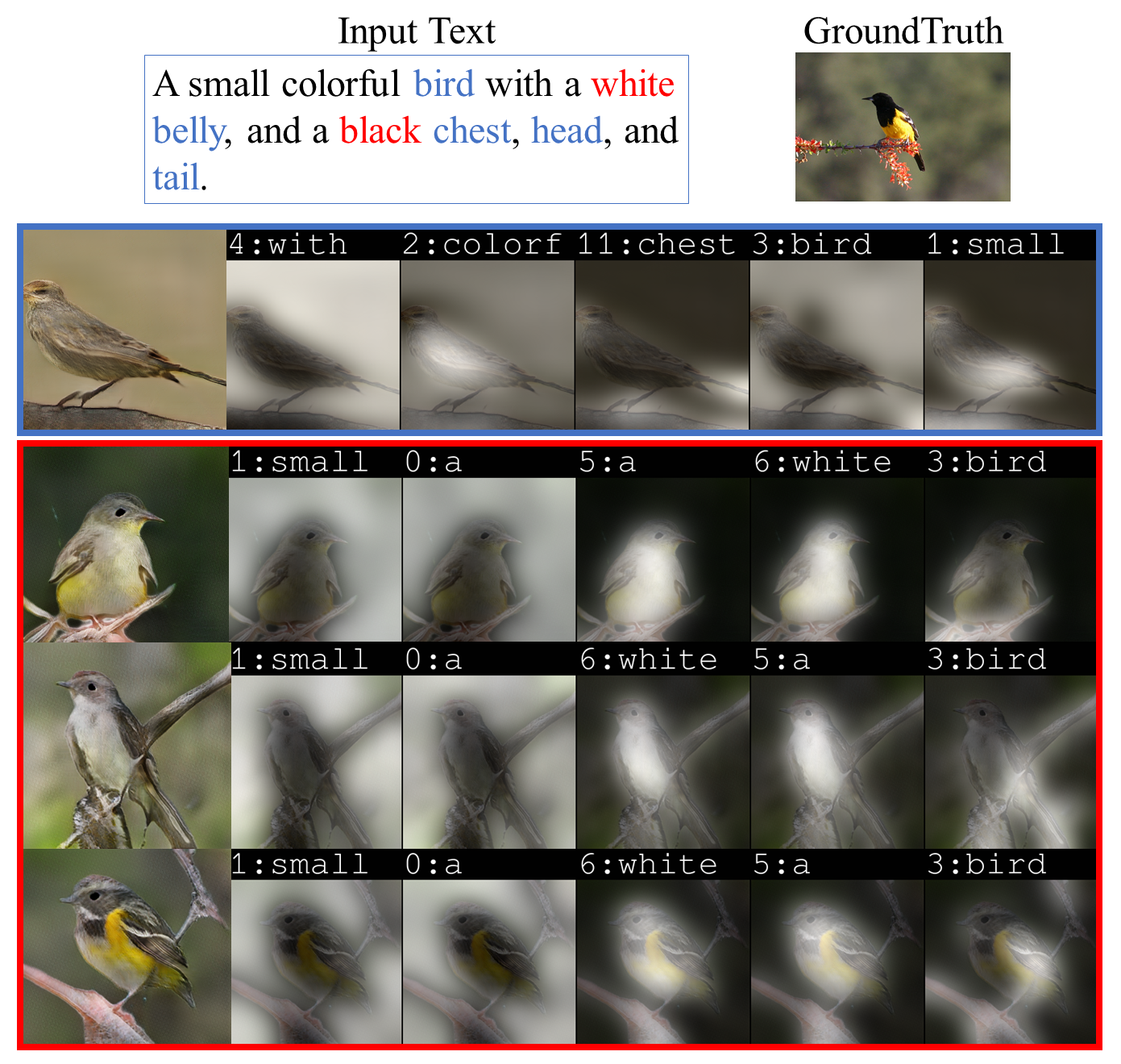}
\end{center}
\caption{$K=3$ synthetic images generated conditioned on the text ``A small colorful bird with a white belly, and a black chest, head, and tail." with the top-5 word attention maps. The results generated by AttnGAN are in blue rectangle, and the results generated by CRD-CGAN are in red rectangle, respectively.}
\label{fig:attnCompare}
\end{figure}

To evaluate the image and text consistency of CRD-CGAN, we visualize the word attention map of CRD-CGAN with AttnGAN in Figure~\ref{fig:attnCompare}, which shows the top-5 word that were attended to by AttnGAN and CRD-CGAN. We can see that the color attribute word {\em``white"} has a lower attention in AttnGAN, while it is the top-5 word in CRD-CGAN.

Compared with StackGAN++, MSGAN and AttnGAN, our proposed CRD-CGAN effectively reduces the FID value in Tabel~\ref{tab:diverse_bird}. At the same time, we also see that the synthetic images generated by D-CGAN-A have a higher FID value than our CRD-CGAN, which means that D-CGAN-A further improves the diversity of synthetic images. This result illustrates that CRD-CGAN reduces the quality of synthetic images compared with D-CGAN-A, this is because the importance of category visual features is emphasized in CRD-CGAN. 
The same results are also appears in Figure~\ref{fig:birds-visual-compare}, where the birds generated by D-CGAN-A have a realistic shape and reasonable color. However, the appearance difference between the generated birds and real birds is still relatively large. On the other hand, the synthetic images generated by CRD-CGAN have the highest scores in user study in Tabel~\ref{tab:diverse_bird} and Tabel~\ref{tab:diverse_flowers}, which means that they have the best shape and color consistency with the corresponding real images.

\begin{table}[h]
\centering
\caption{Inception score comparison on the Birds-200-2011 dataset and the Oxford-102 flower dataset.} 
\begin{tabular}{l|c|c}
\hline
Methods    & CUB                   & Oxford                \\ \hline
StackGAN++&  4.02$\pm $0.58     & 2.49$\pm $0.02              \\
MSGAN&    4.28$\pm $0.05     & 3.25$\pm $0.30  \\
AttnGAN& 4.31$\pm $0.68     & 3.36$\pm $0.02   \\ \hline
D-CGAN-S  &  4.29$\pm $0.07     &3.29$\pm $0.08 \\
D-CGAN-A & 4.51$\pm $0.04 $\uparrow$& 3.39$\pm $0.02 $\uparrow$  \\
RD-CGAN & 4.54$\pm $0.06 $\uparrow$ & 3.48$\pm $0.03$\uparrow$   \\
CD-CGAN & 4.84$\pm $0.11 $\uparrow$    & 3.50$\pm $0.02 $\uparrow$ \\
CRD-CGAN &4.75$\pm $0.10  $\uparrow$ &3.53$\pm $0.06 $\uparrow$  \\
\hline
\end{tabular}
\label{tab:InceptionScore}
\end{table}

To improve the performance of RD-CGAN, we introduce the category consistency loss into D-CGAN-A and RD-CGAN. Firstly, the appearance consistency and diversity of CD-CGAN have been improved compared with D-CGAN-A. The CRD-CGAN also effectively improves the performance of RD-CGAN, and it has the best diversity compared than other methods in Table~\ref{tab:diverse_bird}.The same result is also illustrated in Table~\ref{tab:diverse_flowers}. So we can experimentally confirm that the category consistency loss can effectively constrain the shape and color of the generated images. In other words, the proposed CRD-CGAN can generate synthetic images that are highly realistic and highly consistent with real images. The Inception score~\cite{SalimansGZCRCC16} performance of proposed CRD-CGAN is described in Table~\ref{tab:InceptionScore}, while the StackGAN++, MSGAN and AttnGAN are used for the baselines. From Table~\ref{tab:InceptionScore}, we can confirm the robustness of our work.

To evaluate the correspondence between input text and the synthteic image, we calculate the R-Precision score on the Birds-200-2011 dataset and the Oxford-102 flower dataset. Because of there are $K$ synthetic images, we first calculate the R-Precision score between each synthetic image and the same text correspondingly. Then, we use the mean of $K$ R-Precision scores as the final R-Precision result of $K$ synthetic images. The R-Precision score performance of proposed CRD-CGAN is described in Table~\ref{tab:R-Precision}. From the Table ~\ref{tab:R-Precision}, our CRD-CGAN has the best R-Precision score on the two datasets. In another word, the synthetic images generated by our CRD-CGAN are better able to match the input text.

\begin{table}[h]
\centering
\caption{R-Precision score comparison on the Birds-200-2011 dataset and the Oxford-102 flower dataset.} 
\begin{tabular}{l|c|c}
\hline
Methods    & CUB                   & Oxford                \\ \hline
StackGAN++&  10.57$\pm $4.83     & 13.66$\pm $1.44              \\
MSGAN&    16.08$\pm $5.12   & 18.67$\pm $1.73  \\
AttnGAN& 67.82$\pm $4.43    & 45.50$\pm $1.25   \\ \hline
D-CGAN-S  &  17.33$\pm $4.85     &20.13$\pm $0.98 \\
D-CGAN-A & 68.96$\pm $3.17$\uparrow$ &46.54$\pm $1.56$\uparrow$  \\
RD-CGAN & 70.41$\pm $3.28$\uparrow$ & 56.88$\pm $2.72$\uparrow$   \\
CD-CGAN & 70.62$\pm $2.92$\uparrow$     & 47.12$\pm $1.94$\uparrow$ \\
CRD-CGAN &71.17$\pm $2.36$\uparrow$   &47.70$\pm $2.22$\uparrow$ \\
\hline
\end{tabular}
\label{tab:R-Precision}
\end{table}

\begin{figure*}[h]
\begin{center}
\includegraphics[height=0.68\linewidth, width=0.998\linewidth, angle=0]{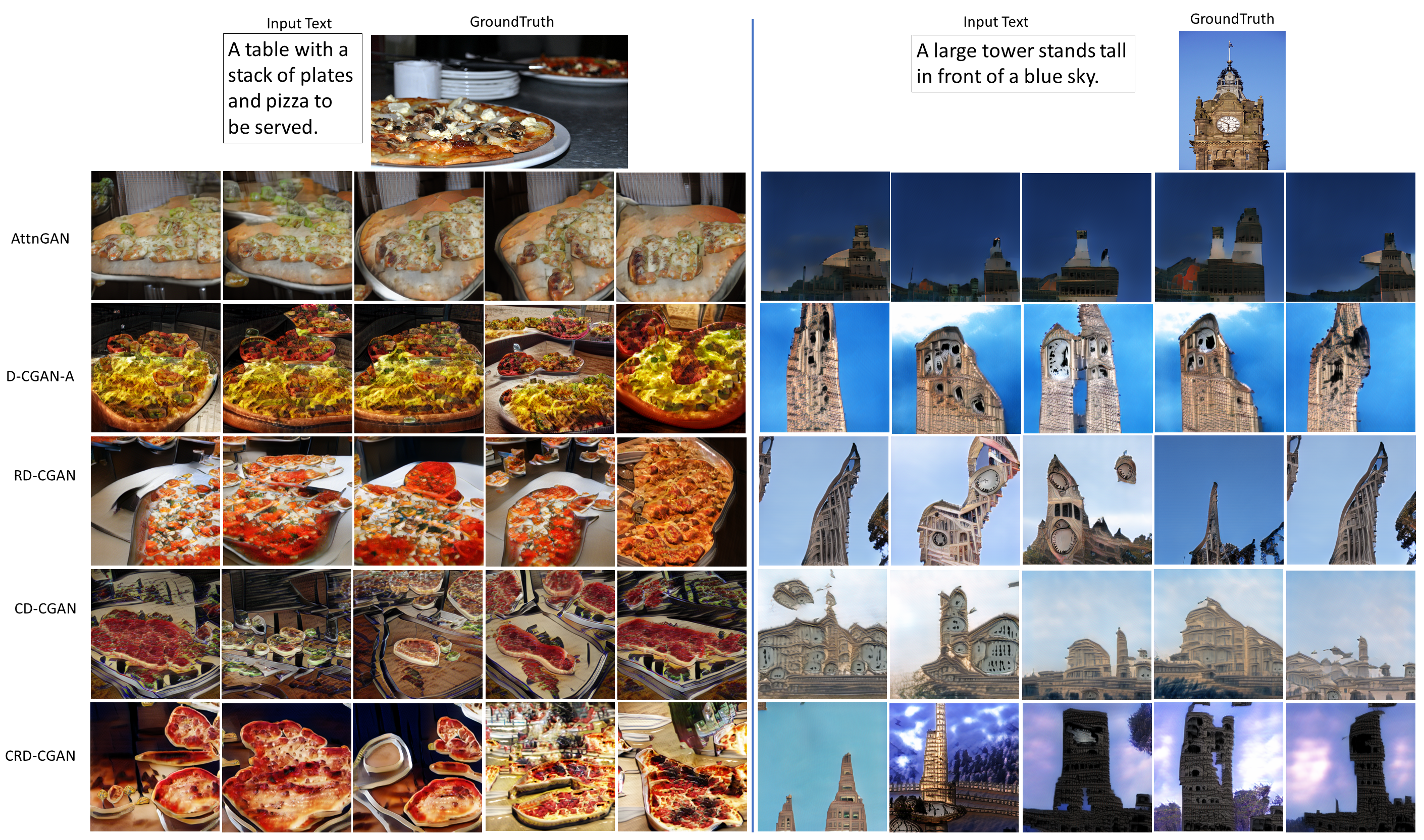}
\end{center}
\caption{Visualization of $K=5$ high-resolution and photo-realistic synthtic images condtioned on a text, and comparing with the corresponding real images (top) on the MS COCO 2014 dataset. }
\label{fig:coco-visual-compare}
\end{figure*}

\subsection{Experiments on the MS COCO 2014 Dataset}

To verify the performance of the proposed CRD-CGAN in complex scenes, we conduct the experiments on the MS COCO 2014 Dataset in this section.

\begin{table}[h]
\centering
\caption{Diversity performance comparison on the MS COCO 2014 dataset.} 
\resizebox{.49\textwidth}{!}{
\begin{tabular}{l|c|c|c}
\hline
Methods    & FID                   & LPIPS  & User study           \\ \hline
AttnGAN
& 42.16$\pm $0.01    &42.06\%$\pm $0.21  &12.41\%$\pm $0.70 \\ \hline
D-CGAN-A & 39.35$\pm $0.02     & 41.49\%$\pm $0.23&16.05\%$\pm $0.23 \\
RD-CGAN &38.61$\pm $0.1 $\downarrow$    & 42.16\%$\pm $0.14&14.64\%$\pm $0.47   \\
CD-CGAN & 43.31$\pm $0.02&42.18\%$\pm $0.43&12.85\%$\pm $0.49 \\
CRD-CGAN &41.79$\pm $0.07&42.52\%$\pm $0.46$\uparrow$&19.45\%$\pm $0.11 \\
\hline
\end{tabular}
\label{tab:diverse_coco}
}
\end{table}

We use the cross entropy to estimate the category probability in Eq.~(\ref{category-loss}), which can calculate the general category of a single object in synthetic image. However, there are muilt-objects in a single image on the MS COCO 2014 Dataset. So we use the multi-class multi-classification hinge loss as a criterion to estimate multi-category probability. We extend the Eq.~(\ref{category-loss}) as follows:
\begin{equation}
\resizebox{.99\linewidth}{!}{$
\displaystyle
\mathcal{L}_{G_i}^{CC}=
-\sum_{y=1}^{Y}\frac{\max(0, 1 - \textit{log}P(y|X_i, s_i^1, \ldots, s_i^K) - \delta(y\in {\bf c}_i)))}{|{\bf c}_i|}
$}
\label{multi-category-loss}
\end{equation}
where $Y$ is the total category number of COCO  dataset, $\delta(y \in {\bf c}_i)$ is 1 when $y$ is one of the true multi-labels ${\bf c}_i$ of the real image $X_i$ and 0 otherwise,  and $|{\bf c}_i|$ is the total number of multi-labels for $X_i$. 

\begin{figure*}[h]
\begin{center}
\includegraphics[height=0.999\linewidth, width=0.998\linewidth, angle=0]{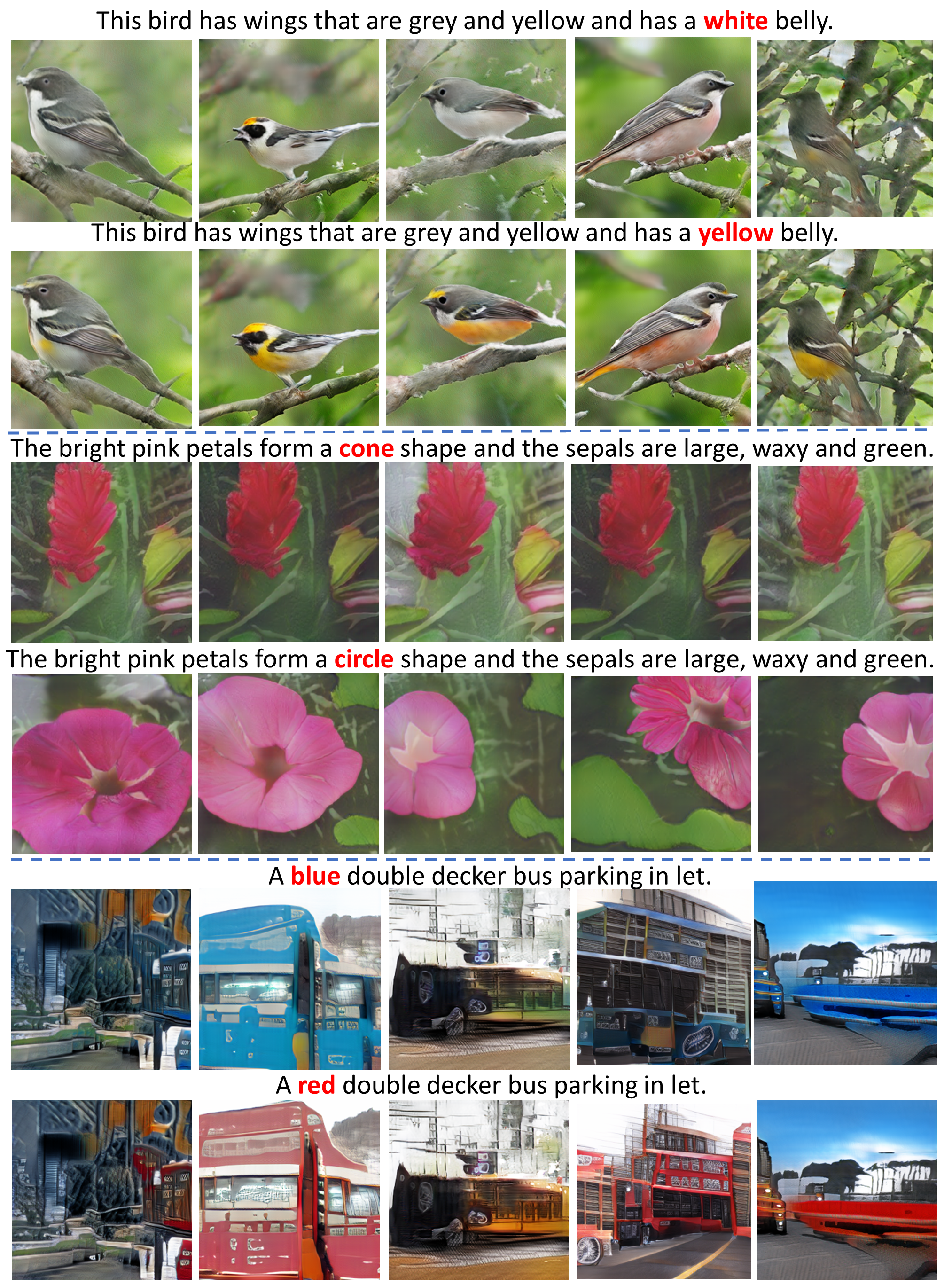}
\end{center}
\caption{Examples of CRD-CGAN on the ability of catching words changes (underline word in red) of the text description on the Birds-200-2011 dataset (top), on the Oxford-102 flower dataset (middle), and on MS COCO 2014 dataset(below).}
\label{fig:change}
\end{figure*}

We use two Nvidia Titan RTX GPUs to train the proposed CRD-CGAN on the MS COCO 2014 dataset, and the GPU memory is up to 48GB. Due to complex loss calculations and large-scale dataset, we set $K=5$ to train the proposed RD-CGAN, CD-CGAN and CRD-CGAN. The FID and LPIPS scores for our proposed CRDCGAN and other methods on the MS COCO 2014 dataset are summarized in Table~\ref{tab:diverse_coco}. From Table~\ref{tab:diverse_coco}, we can see the quality of synthetic images generated by D-CGAN-A, RD-CGAN, CD-CGAN and CRD-CGAN are higher than those by AttnGAN.The synthetic images generated by CRD-CGAN exhibit the highest diversity than other methods. Our CRD-CGAN achieves the best score in the user study of similarity comparison, which also shows that the synthetic images generated by CRD-CGAN is the most similar to the real images. 

We also visualize some synthetic images by our CRD-CGAN and other methods in Figure~\ref{fig:coco-visual-compare}. We can observe the synthetic images generated by AttnGAN have less diversity and have lower similarity to real images. While the sythetic images generated by our RD-CGAN, CD-CGAN and CRD-CGAN have better shape in details. For example, the input text is {\em``A table with a stack of plates and pizza to be served"}. The words {\em``table"}, {\em``plates"} and {\em``pizza"} are the main word, which are reflected in the synthetic images generated by our CRD-CGAN. In contrast, the results of AttnGAN only show a simple shape. 

\subsection{Semantic Sensitivity Application}
Furthermore, to evaluate the semantic sensitivity of the proposed CRD-CGAN, we change just one word in the input text. As shown in Figure~\ref{fig:change}, the synthetic images are modified according to the changes of the input texts. For example, the color of bird ``white" is changed to ``yellow", the shape of flower ``cone" is changed to ``circle", and the color of bus ``blue" is changed to ``red". It demonstrates the proposed CRD-CGAN has the ability to retain the semantic diversity by catching the changes of the text description.
%

\section{Conclusion}
In this paper, we employed the category-consistent and relativistic diverse constraints to effectively exploit the relative real-or-fake relationship and main visual consistency between real image and $K$ synthetic images. Our CRD-CGAN improves the estimating probability of more realism or more artifacts of $K$ synthetic images relative to real image, and it uses the category consistency loss to ensure that the $K$ synthetic images retain the main visual feature of corresponding category. Extensive experiments demonstrate the respected effectiveness and significance of proposed CRD-CGAN on the Birds-200-211 and Oxford-102 flower dataset. To evaluate the performance on lager-scale dataset, we also test our method on MS COCO 2014 datasets. The experiments results show that the proposed CRD-CGAN is also applicable to generate complex scenes with multiple categories.

{\small
\bibliographystyle{ieee_fullname}
\bibliography{CRDGAN}
}

\end{document}